\documentclass{article}

    \usepackage{geometry}
    \usepackage{textcomp}
    \usepackage{adjustbox}
    \usepackage{mathtools}
    \usepackage{booktabs} %
    \usepackage[group-separator={,},group-minimum-digits=4]{siunitx}
    \usepackage{changepage}
\usepackage{arxiv}
\usepackage[T1]{fontenc}    
\usepackage{hyperref}       
\usepackage{url}            
\usepackage{booktabs}       
\usepackage{amsfonts}       
\usepackage{nicefrac}       
\usepackage{microtype}      
\usepackage{lipsum}
\usepackage[english]{babel}
\usepackage{optidef}
\usepackage[nottoc]{tocbibind}
\usepackage{amssymb}
\usepackage[square,numbers]{natbib}
\usepackage[ruled,vlined]{algorithm2e}
\newcommand{\norm}[1]{\| #1 \|}
\usepackage{graphicx}
\graphicspath{ {./plots/} }
\usepackage[toc,page]{appendix}
\usepackage{array}
\usepackage{tikz}
\usetikzlibrary{decorations.pathreplacing}
\usepackage{multirow}
\usepackage{comment}

\usepackage{amsmath}
\usepackage{amsmath,amsfonts,amssymb}
\usepackage{amsmath,amssymb}
\usepackage{booktabs}
\usepackage{multirow}
\usepackage{siunitx}

\title{Neuron-based Pruning of Deep Neural Networks with Better Generalization using Kronecker Factored Curvature Approximation}

\author{
  Abdolghani~Ebrahimi\\
  Department of Industrial Engineering \& Management Sciences\\
  Northwestern University\\
  Evanston, IL 60208 \\
  \texttt{ebrahimi@u.northwestern.edu} \\
   \And
 Diego~Klabjan \\
  Department of Industrial Engineering \& Management Sciences\\
  Northwestern University\\
  Evanston, IL 60208 \\
  \texttt{d-klabjan@northwestern.edu} \\
}
\usepackage{indentfirst}
\begin{document}
\maketitle
\begin{abstract}
\par Existing methods of pruning deep neural networks focus on removing unnecessary parameters of the trained network and fine tuning the model afterwards to find a good solution that recovers the initial performance of the trained model. Unlike other works, our method pays special attention to the quality of the solution in the compressed model and inference computation time by pruning neurons. The proposed algorithm directs the parameters of the compressed model toward a flatter solution by exploring the spectral radius of Hessian which results in better generalization on unseen data. Moreover, the method does not work with a pre-trained network and performs training and pruning simultaneously. Our result shows that it improves the state-of-the-art results on neuron compression. The method is able to achieve very small networks with small accuracy degradation across different neural network models.   
\end{abstract}


\section{Introduction} \label{introduction}
Deep neural networks (DNNs) are now widely used in areas like object detection and natural language processing due to their unprecedented success. It is known that the performance of DNNs often improves by increasing the number of layers and neurons per layer. Training and deploying these deep networks sometimes call for devices with excessive computational power and high memory. In this situation, the idea of pruning parameters of DNNs with minimal drop in the performance is crucial for real-time applications on devices such as cellular phones with limited computational and memory resources. In this paper, we propose an algorithm for pruning neurons in DNNs with special attention to the performance of the pruned network. In other words, our algorithm directs the parameters of the pruned network toward flatter areas and therefore better generalization on unseen data.

In all network pruning studies, the goal is to learn a network with a much smaller number of parameters which is able to virtually match the performance of the overparameterized network. Most of the works in this area have focused on pruning the weights in the network rather than pruning neurons. As \cite{molchanov2016pruning} and \cite{singh2019play} discuss in their work, pruning weights by setting them to zero does not necessarily translate to a faster inference time since specialized software designed to work with sparse tensors is needed to achieve this. Our work, however, is focused on pruning neurons from the network which directly translates to a faster inference time using current regular software since neuron pruning results in lower dimensional tensors. 

The pruning framework proposed herein iterates the following two steps: (1) given a subset of neurons, we train the underlying network with the aim of low loss and “flatness,” (2) given a trained network on a subset of neurons, we compute the gradient of each neuron with respect to being present or not in the next subset of neurons. Regarding the first step, we consider flatness as being measured by the spectral radius of Hessian. This requires computing the spectral radius which is approximated by computing the spectral radius of its Kronecker-factored Approximate Curvature (K-FAC) block diagonal approximation. The flatness concept comes from \cite{adam} but using K-FAC is brand new. In the second step, the selection of a neuron is binary which is approximated as continuous in the interval [0,1]. The gradient is then computed with respect to loss plus spectral radius of K-FAC. Neurons with large gradient values are selected for the next network.  

The methodology proposed improves the generalization accuracy results on validation datasets in comparison to the state-of-the-art results. In all of the four experiments on different architectures, our results outperform when the sparsity level is lower than a threshold value (around $50\%$ sparsity level for three datasets). The area under the curve with respect to sparsity and accuracy is higher for our approach from around $1\%$ and up to $9\%$.
    
There is a vast body of work on network pruning and compression (see \cite{cheng2017survey} for a full list of works), however, their focus is devising algorithms to produce smaller networks without a substantial drop in accuracy. None of those works tries to find solutions with flatter minima in the compressed network. To the best of our knowledge, our work is the first study that takes that into account and tries to find solutions with a better generalization property.  Specifically, we make the following contributions.
    \begin{itemize}
        \item We build upon the work of \cite{adam} and improve their work by using K-FAC to compute the spectral radius of Hessian and its corresponding eigenvector. Our algorithm can be easily parallelized for faster computation. This avoids using a power iteration algorithm that might not converge in a reasonable number of iterations and time. 
        \item We provide an algorithm for learning a small network with a low spectral radius from a much bigger network. Despite pruning a big portion of the neurons, the accuracy remains almost the same as in the bigger network across different architectures.
        \item Our method allows for aggressive pruning of neurons of the neural network in each pruning epoch, whereas other methods such as \cite{molchanov2016pruning} are conservative and prune just one neuron per pruning epoch.
        \item Our algorithm is able to achieve very small networks with really small accuracy degradation across different neural network models and outperforms existing literature on neuron pruning across most of the network architectures we experiment with. 
    \end{itemize}
    
    The rest of this paper is organized as follows. In Section \ref{sec:lr}, we overview the related works. We formally define our problem, model it as an optimization problem and provide an algorithm to solve it in Section \ref{sec:model_algorithm}. Finally, In Section \ref{sec:computation}, we apply the algorithm across different modern architectures and conduct an ablation study to demonstrate effectiveness of the method and discuss the results.   

    \section{Literature Review}
        \label{sec:lr}
        

        Our work is closely related to two different streams of research: 1) network pruning and compression; and 2) better generalization in deep learning. We review each of the two streams in the following paragraphs.
        
        \textbf{Network pruning and compression.} \quad Study \cite{denil2013predicting} demonstrates significant redundancy of parameters across several different architectures. This redundancy results in waste of computation and memory and often culminates in overfitting. This opens up a huge opportunity to develop ideas to shrink overparameterized networks. Most of the works in the area focus on pruning weights. Earlier works on weight pruning started with Optimal Brain Damage \cite{lecun1990optimal} and Optimal Brain Surgeon \cite{hassibi1993second} algorithms. These methods focus on the Taylor expansion of loss functions and prune those weights which increase the loss function by the smallest value. They need to calculate Hessian of the loss function which could be cumbersome in big networks with millions of parameters. To avoid this, they use a diagonal approximation of Hessian. Regularization is another common way for pruning DNNs and coping with overfitting. \cite{ishikawa1996structural} and \cite{chauvin1989back} augment the loss function with $L_0$ or $L_1$ regularization terms. That sets some of the weights very close to zero. Eventually, one can remove the weights with value smaller than a threshold. \cite{han2015learning} is one of the most successful works in weight pruning. They simply use a trained model and drop the weights with the value below a threshold and fine-tune the resulting network. This procedure can be repeated until a specific sparsity level is achieved. However, this alternation between pruning and training can be long and tedious. To avoid this, \cite{lee2018snip} develops an algorithm that prunes the weights before training the network. They use a mask variable for each weight and calculate the derivative of the loss function with respect to those mask variables and use that as a proxy for sensitivity of the the loss function to that weight. Our  pruning scheme is somewhat related to this work. We consider sensitivity of the loss function including spectral radius with respect to neurons rather than weights.
        
On the other hand, some compression works focus on pruning neurons. Our work falls into this group of methods. As \cite{molchanov2016pruning} discusses, given current software, pruning neurons is more effective to result in a faster inference time and that is our main reason to focus on pruning neurons in this paper. \cite{srinivas2015data} uses a data-free pruning method based on the idea of similar neurons. Their method is only applicable to fully connected networks. \cite{hu2016network} notices that most of neurons in large networks have activation values close to zero regardless of the input to those neurons and do not play a substantial part in prediction accuracy. This observation leads to a scheme to prune these zero-activation neurons. To recover the performance of the network from pruning these neurons they alternate between pruning neurons and training. Some of the works on neuron pruning are specifically designed for Convolutional Neural Networks (CNNs) as they are much more computationally demanding compared to fully connected layers \cite{li2016pruning}. There is abundant work on pruning filters (feature maps) in CNNs (see \cite{singh2019play, li2016pruning, anwar2017structured} as examples). Our neuron pruning scheme has been inspired by the work of \cite{molchanov2016pruning} on CNN filter pruning. They adopt a Taylor expansion of the loss function to find how much pruning each filter in a convolution layer changes the loss function and prune those that contribute the least to the change. Although, we use the same pruning criteria, our work is different from theirs in the following three substantial ways. First, they do work with fully trained models and apply pruning to it. Instead, our work integrates the training and pruning parts together. Second, we augment the loss function with a term related to the spectral radius of Hessian and thus our neuron pruning scheme is sensitive to the change in flatness of a solution in addition to the change in the loss function. Finally, our pruning scheme is not just applied to prune filters in CNNs. It is more general and can be used to prune other type of layers such as fully connected layers.


\textbf{Better generalization in deep learning.}\quad The initial work \cite{lecun2012efficient} observes that the Stochastic Gradient Descent (SGD) algorithm and similar methods such as RMSProp \cite{hinton2012neural} and ADAM \cite{kingma2014adam} generalize better on new data when the network is trained with smaller mini-batch sizes. \cite{keskar2016large} experimentally demonstrates that using a smaller mini-batch size with SGD tends to converge to flatter minima and that is the main reason for better generalization. \cite{jastrzebski2018finding} later shows that using larger learning rates also plays a key role in converging to flatter minima. Their work essentially studies the effect of the learning rate to the mini-batch size ratio on the generalization of found solutions on unseen data. Almost all works in this area focus on investigating the effect of hyperparameters such as the mini-batch size and learning rate on the curvature of the achieved solution. Despite this finding, there has been little effort on devising algorithms that are guaranteed to converge to flatter minima. Study \cite{adam} is the only work that explores this problem in deep learning settings and tries to find a flatter minimum solution algorithmically. To achieve flatter solutions, they augment the loss function with the spectral radius of Hessian of the loss function. To compute the latter, they use power iteration along with an efficient procedure known as R-Op \cite{pearlmutter1994fast} to calculate the Hessian vector product. This implies that they need to use the power iteration algorithm in every optimization step of their algorithm. One potential drawback of using power iteration is that it might not converge in a reasonable number of iterations. To avoid this, we borrow ideas from K-FAC which is originally developed by \cite{martens2015optimizing} and \cite{grosse2016kronecker} to apply natural gradient descent \cite{amari1998natural} to deep learning optimization. Our proposed method adds the same penalty term, however we use K-FAC as a different approach to find an approximate spectral radius of Hessian. In addition, we use the augmented loss function not only for training but also for the pruning purpose. K-FAC uses the Kronecker product to approximate the Fisher information matrix which is a positive semi-definite approximation of Hessian. K-FAC approximates the Fisher information matrix by a block diagonal matrix where each block is associated with a layer in the neural network architecture. Another positive semi-definite approximation that is widely used instead of Hessian is the Gauss-Newton (GN) matrix. Work \cite{botev2017practical} shows that when activation functions used in the neural network are piecewise linear, e.g., standard rectified linear unit (ReLU) and leaky rectified linear unit (Leaky ReLU), the diagonal blocks in Hessian and GN are the same. On the other hand, \cite{martens2014new} and \cite{pascanu2013revisiting} prove that for typical loss functions in machine learning such as cross-entropy and squared error, the Fisher information and the GN matrices are the same. These observations justify using K-FAC to approximate Hessian of each layer in a deep learning setting as most of the state-of-the-art neural network architectures use ReLU as their activation function.

       \section{Proposed Approach}
        \label{sec:model_algorithm}
        Our goal here is designing an algorithm that given a data set and a neural network architecture carefully prunes redundant neurons during training and produces a smaller network with a flat minimum solution. To this end, we first introduce the underlying optimization problem and then discuss the algorithm to solve it.

        \subsection{Optimization Problem for Integrated Pruning and Training}
We are given training data $\mathcal{D} = \{(\mathbf{x}_i, y_i)\}_{i = 1}^{n}$ where $\mathbf{x}$ represents features and $y$ represents target values and a neural network with a total of $\mathcal{N}$ neurons. We define the set of all the parameters of the neural network with $\ell$ layers and the total number of parameters $d$ as $\mathcal{W} = (\mathbf{w}_1^1, \mathbf{w}_1^2, ..., \mathbf{w}_1^{n_1} , ..., \mathbf{w}_\ell ^{1}, \mathbf{w}_\ell ^{2}, ..., \mathbf{w}_\ell ^{n_\ell}) \in \mathbb{R}^d$. Here, $n_l$ represents the number of neurons in layer $l\in [\ell]$ (where $[x]:=\{1, 2, ..., x\}$) and $\mathbf{w}_l^j$ represents the parameters associated with neuron $j \in [n_\ell]$ in layer $l \in [\ell]$. 
Since we need the optimization problem to identify and prune redundant neurons, we introduce the set of binary (mask) variables $\mathcal{M_B} = (m_1^1, m_1^2, ..., m_1^{n_1} , ..., m_\ell ^{1}, m_\ell ^{2}, ..., m_\ell ^{n_\ell}) \in \{0, 1\}^\mathcal{N}$. We also define $\mathcal{M} = (m_1^1 e_{\alpha(1,1)}, m_1^2e_{\alpha(1,2)}, ..., m_1^{n_1}e_{\alpha(1,n_1)} , ..., m_\ell ^{1}e_{\alpha(\ell,1)}, m_\ell ^{2}e_{\alpha(\ell,2)}, ..., m_\ell ^{n_\ell}e_{\alpha(\ell,n_\ell)}) \in \{0, 1\}^d$ where $e_s = (1, ..., 1) \in \mathbb{R}^s$ and $\alpha(i,j)$ is the dimension of $\mathbf{w}_i^j$. Each element in $\mathcal{M}_B$ is associated with one of the neurons and identifies whether a neuron fires (value 1) or needs to be pruned (value 0) in the neural network architecture.  To achieve a solution with the flatter curvature, we represent the spectral radius of Hessian (or its approximation) of the loss function $\mathcal{C}$  (e.g. cross entropy) at point $\mathcal{W}$ as $\rho_\mathcal{C}(H(\mathcal{W}))$. Then, given the maximum allowed number of neurons $K$ and upper bound $B$ on the spectral radius of Hessian, the optimization problem can be written as the following constrained problem:
        \begin{equation}
        \label{dfn:our_model1}
        \begin{aligned}
        \min_{\mathcal{W},\mathcal{M}_B} \ & \mathcal{C}(\mathcal{W} \odot \mathcal{M};\mathcal{D}), \\
        \text{s.t. }\  
        & \rho_{\mathcal{C}}(H(\mathcal{\mathcal{W} \odot \mathcal{M}})) \leq B,\\
        & \mathcal{M}^T_B\mathbf{e}_\mathcal{N}  \leq K,
        \end{aligned}
        \end{equation}
where $\odot$ is the Hadamard product. One drawback of formulation (\ref{dfn:our_model1}) is intractability of a direct derivation of Hessian of the loss function and subsequently its spectral radius in DNNs with millions of parameters. To cope with this, we follow \cite{adam} and rewrite $\rho(H(\mathcal{\mathcal{W} \odot \mathcal{M}}))$ in terms of vector $v(\mathcal{W},\mathcal{M})$ where $\norm{v(\mathcal{W},\mathcal{M})} = 1$ and it denotes the eigenvector corresponding to the eigenvalue of $H(\mathcal{\mathcal{W} \odot \mathcal{M}})$ with the largest absolute value. We also introduce the constraint on the spectral radius to the objective function as follows:
        \begin{equation}
        \label{dfn:our_model2}
        \begin{aligned}
        \min_{\mathcal{W},\mathcal{M}_B} \ & L(\mathcal{W}, \mathcal{M}_B; \mathcal{D}) = \mathcal{C}(\mathcal{W} \odot \mathcal{M};\mathcal{D}) + \mu g(\mathcal{W},\mathcal{M}) \\
        \text{s.t. }\  
        & \mathcal{M}_B^T\mathbf{e}_\mathcal{N}  \leq K,
        \end{aligned}
        \end{equation}
        where $g(\mathcal{W},\mathcal{M}) := \max\{0, \mathbf{v}^T(\mathcal{W},\mathcal{M})H(\mathcal{\mathcal{W} \odot \mathcal{M}})\mathbf{v}(\mathcal{W},\mathcal{M}) - B\}$ and $\mu$ is a hyperparameter which needs to be tuned. The larger the $\mu$ is, the optimization problem penalizes larger spectral radii more and the loss function does have a flatter curvature at the solution point. To calculate $g(\mathcal{W},\mathcal{M})$, one needs to be able to calculate $\mathbf{v}(\mathcal{W},\mathcal{M})$. To this end, we follow \cite{martens2015optimizing} and approximate the Hessian of the loss function with a block diagonal matrix where each block is the second order derivative of the loss function with respect to the parameters of each layer. In other words, denoting activations and inputs (a.k.a pre-activations) for layer $l \in [\ell]$ with $\mathbf{a}_l(\mathcal{W}, \mathcal{M})$ and $\mathbf{s}_l(\mathcal{W}, \mathcal{M})$, respectively, and defining $\mathbf{g}_l(\mathcal{W}, \mathcal{M}):= \frac{\partial L(\mathcal{W}, \mathcal{M}; \mathcal{D})}{\partial \mathbf{s}_l(\mathcal{W}, \mathcal{M})}$ we have:
\begin{equation} 
\label{hessmatrix}
H(\mathcal{W} \odot \mathcal{M}) \approx 
    \left(
    \begin{array}{ccccc}
    \mathbf{\Psi}_0(\mathcal{W} \odot \mathcal{M})\otimes\mathbf{\Gamma}_1(\mathcal{W} \odot \mathcal{M}) & & \textbf{0}\\
      & \ddots \\
      \textbf{0}  &             & \mathbf{\Psi}_{\ell-1}(\mathcal{W} \odot \mathcal{M})\otimes\mathbf{\Gamma}_\ell(\mathcal{W} \odot \mathcal{M})
    \end{array}
    \right).
\end{equation}
In (\ref{hessmatrix}), $\mathbf{\Psi}_l(\mathcal{W} \odot \mathcal{M}) := \mathbb{E}[\mathbf{a}_l(\mathcal{W}, \mathcal{M})\mathbf{a}_l^T(\mathcal{W}, \mathcal{M})]$ and $\mathbf{\Gamma}_l(\mathcal{W} \odot \mathcal{M}) := \mathbb{E}[\mathbf{g}_l(\mathcal{W}, \mathcal{M})\mathbf{g}_l^T(\mathcal{W}, \mathcal{M})]$ denote matrices of the second moment for activations and derivative of pre-activations for layer $l$, respectively. Notice that no additional computational resources are needed to derive $\mathbf{g}_l(\mathcal{W}, \mathcal{M})$ and $\mathbf{a}_l(\mathcal{W}, \mathcal{M})$ as they are derived as a byproduct of the back-propagation procedure.
Let $\mathbf{v}_l(\mathcal{W}, \mathcal{M})$ represent the eigenvector corresponding to the largest absolute eigenvalue $\lambda_l(\mathcal{W}, \mathcal{M})$ for block (layer) $l$. We also define the same eigenvalues $
\lambda^\Psi_l(\mathcal{W}, \mathcal{M})$ and $
\lambda^\Gamma_l(\mathcal{W}, \mathcal{M})$ along with their corresponding eigenvectors $
\mathbf{v}^\Psi_l(\mathcal{W}, \mathcal{M})$ and $
\mathbf{v}^\Gamma_l(\mathcal{W}, \mathcal{M})$ for matrices $\mathbf{\Psi}_{l}(\mathcal{W} \odot \mathcal{M})$ and $\mathbf{\Gamma}_l(\mathcal{W} \odot \mathcal{M})$, respectively. Then, we can use properties of the Kronecker product (see \cite{schacke2004kronecker}) to produce:
\begin{equation}
\label{model:kfac}
\begin{aligned} 
\mathbf{v}_l(\mathcal{W}, \mathcal{M}) &= \mathbf{v}^\Psi_{l-1}(\mathcal{W}, \mathcal{M}) \otimes \mathbf{v}^\Gamma_l(\mathcal{W}, \mathcal{M}) \\ 
\lambda_l(\mathcal{W}, \mathcal{M}) &=  \lambda^\Psi_{l-1}(\mathcal{W}, \mathcal{M}) \lambda^\Gamma_l(\mathcal{W}, \mathcal{M}).
\end{aligned}   
\end{equation}

Once we have $\mathbf{v}_l(\mathcal{W}, \mathcal{M})$ for each block $l$, approximating $\mathbf{v}(\mathcal{W},\mathcal{M})$ is trivial given the properties of block diagonal matrices.

Optimization problem (\ref{dfn:our_model2}) has $\mathcal{N}$ additional variables $\mathcal{M}_B$ and is not directly solvable using continuous optimization methods like SGD due to the existence of the non-continuous mask variables $\mathcal{M}_B$. However, the separation of each neuron parameters
from neuron masks, facilitates measuring the contribution of each of the neurons to the loss function. In other words, we consider the sensitivity of the loss function to each neuron as a measure for importance and only keep the neurons with highest sensitivity. To this end, since variables in $\mathcal{M}_B$ are binary, sensitivity of the loss function with respect to neuron $j \in [n_\ell]$ in layer $l \in [\ell]$ is defined as:
        \begin{equation}
        \label{dfn:our_model3}
        \Delta  \bar{L}_l^j(\mathcal{W}, \mathcal{M}_B) := \: \mid L(\mathcal{W}, \mathcal{M}_B| m_l^j = 0 ;  \mathcal{D}) - L(\mathcal{W}, \mathcal{M}_B| m_l^j = 1 ; \mathcal{D})\mid,
        \end{equation}
where $\mathcal{M}_B| m_l^j = 0$, $\mathcal{M}_B| m_l^j = 1$ are two vectors with same values for all neurons except for the neuron $j$ in layer $l$. Following \cite{molchanov2016pruning} and using first order Taylor approximation of $L(\mathcal{W}, \mathcal{M}_B| m_l^j = 0)$
around $\mathbf{a}_l^j(\mathcal{W}, \mathcal{M}) = 0$, we estimate (\ref{dfn:our_model3}) by

        \begin{equation}
        \label{dfn:equation}
        \Delta  \bar{L} _l^j(\mathcal{W}, \mathcal{M}_B) := \: \mid \frac{\partial L(\mathcal{W}, \mathcal{M}_B; \mathcal{D})}{\partial \mathbf{a}_l^j(\mathcal{W}, \mathcal{M})}\mathbf{a}_l^j(\mathcal{W}, \mathcal{M}) \mid.
        \end{equation}
This intuitively means that the importance of a neuron in the neural network is dependent on two values: the magnitude of the gradient of the loss function with respect to the neuron's activation and the magnitude of the activation. It is worth mentioning that the numeric value of  $\frac{\partial L(\mathcal{W}, \mathcal{M}_B; \mathcal{D})}{\partial \mathbf{a}_l^j(\mathcal{W}, \mathcal{M})}$ is also derived during the back-propagation procedure and no further computation is necessary. Following our discussion from above, we provide the algorithm to solve minimization problem (\ref{dfn:our_model2}). 

\subsection{Algorithm}
\label{Algorithm}
The algorithm, Higher Generalization Neuron Pruning (HGNP), solves optimization problem \ref{dfn:our_model2} by updating $\mathcal{W}$ and $\mathcal{M}$ alternately. When the parameters $\mathcal{W}$ are being updated, we fix the binary mask vector $\mathcal{M}$ and vice versa. The alternation procedure between updating mask vector $\mathcal{M}$ and weights $\mathcal{W}$ happens gradually during training to avoid a substantial drop in the performance of the model. Pruning an excessive number of neurons in one shot might result in an irrecoverable performance drop and might result in low validation accuracy of the final pruned model. To update the parameters $\mathcal{W}$, we fix the mask variables $\mathcal{M}_B = \hat{\mathcal{M}}_B$ for a certain number of iterations and update weights $\mathcal{W}$ by following Algorithm \ref{algo1}. This algorithm essentially solves the optimization problem $min_\mathcal{W} L(\mathcal{W}, \hat{\mathcal{M}_B}; \mathcal{D})$ for fixed $\hat{\mathcal{M}_B}$.
\begin{algorithm}[H]
\textbf{Input:} $\mathcal{D}$ : training data, $\hat{\mathcal{M}}_B$: given mask vector, $\hat{\mathcal{W}}$: given initialization weights, $\alpha$ : learning rate, $\mu$ : coefficient for spectral radius, $B$: bound on spectral radius.\\
\textbf{Output:} $\mathcal{W}$ : the updated weights.\\ 
Initialize $\mathcal{W} \leftarrow \hat{\mathcal{W}}$.\\
\While{convergence criteria not met}{
\For{each batch of data  $\mathcal{D}^b$}
{
Compute $\nabla_\mathcal{W} \mathcal{C} (\mathcal{W}\odot \hat{\mathcal{M}} ; \mathcal{D}^b)$\\
\tcc{Spectral radius term operations}
\For{$l = 0, 1, ..., \ell$}
{
Derive $\mathbf{v}_l^\psi(\mathcal{W}, \hat{\mathcal{M}})$, $\mathbf{v}_l^\gamma(\mathcal{W}, \hat{\mathcal{M}})$ and eigenvalues $\lambda_l^\psi(\mathcal{W}, \hat{\mathcal{M}})$, $\lambda_l^\gamma(\mathcal{W}, \hat{\mathcal{M}})$.\\
Compute $\mathbf{v}_l(\mathcal{W}, \hat{\mathcal{M}})$ and $\lambda_l(\mathcal{W}, \hat{\mathcal{M}})$ using \ref{model:kfac}.}
Approximate $v(\mathcal{W}, \hat{\mathcal{M}})$ and its corresponding eigenvalue $\rho_\mathcal{C}(H(\mathcal{W} \odot \hat{\mathcal{M}})$ using eigenvectors and eigenvalues derived for each block.\\
Compute using R$^2$-Op from \cite{adam}:
$\nabla_\mathcal{W}\rho_\mathcal{C}(H(\mathcal{W} \odot \hat{\mathcal{M}})=\frac{1}{|\mathcal{D}^b|}\sum\limits_{i\in \mathcal{D}^b} v(\mathcal{W}, \hat{\mathcal{M}})^T\nabla_\mathcal{W} H_i(\mathcal{W} \odot \hat{\mathcal{M}}) v(\mathcal{W}, \hat{\mathcal{M}})$ where index $i$ shows that Hessian is computed with respect to single sample $i$.\\
Derive $\nabla_\mathcal{W} g(\mathcal{W}, \hat{\mathcal{M}})$ using $\nabla_
\mathcal{W}\rho_\mathcal{C}(H(\mathcal{W} \odot \hat{\mathcal{M}})$.\\
\tcc{Gradient descent}
Set $\mathcal{W}=\mathcal{W}-\alpha \left(\nabla_\mathcal{W} \mathcal{C} (\mathcal{W}\odot \hat{\mathcal{M}} ; \mathcal{D}^b) +\mu \nabla_\mathcal{W} g(\mathcal{W}, \hat{\mathcal{M}})\right)$
}}
\caption{Mini-Batch SGD Algorithm to update $\mathcal{W}$ for fixed mask vector $\hat{\mathcal{M}}_B$}
\label{algo1}
\end{algorithm}
 To update $\mathcal{M}_B$, we apply \ref{dfn:equation} to a random mini-batch to find the neurons that the loss function is least sensitive to. We update $\mathcal{M}_B$ by setting the mask variable associated with such neurons to zero. It means that we can remove such neurons from the neural network and recreate a new neural network with the remaining neurons and their associated weights in the parameters vector $\mathcal{W}$. We now can use this concept to prune neurons along with Algorithm \ref{algo1} to propose Algorithm \ref{main_algo}.
 
The algorithm starts with random initialization of parameters $\mathcal{W}_0$. We fix the mask variables $\mathcal{M}_B$ to all one at first and update $\mathcal{W}$ using SGD for $E_1$ training epochs. After taking the pre-pruning training steps, we prune the neural network for the first time by pruning the $N$ neurons with the smallest effect on the loss function change. We then alternate between training and pruning each $E_2$ epochs and continue pruning until a specific sparsity level $\bar{\kappa}$ is achieved. Here, we define sparsity level $\kappa := \frac{\|\mathcal{W} \odot \mathcal{M}\|_0}{\|\mathcal{W} \odot e_d\|_0}$. Although we stop the algorithm using the same metric, i.e. sparsity level, like other works in weight pruning, there is a fundamental difference between the actual network architecture after pruning in our work and other works. In our case, we remove all the weights associated with the pruned neurons, but in other works, such weights are set to zero. Once the desired sparsity level $\bar{\kappa}$ is achieved, i.e. $\kappa \leq \bar{\kappa}$, we train the network for the last $E_3$ epochs. One should notice that values $E_1$, $E_2$, $E_3$ and $N$ are hyperparameters and need to be tuned and vary across different neural network architectures. We summarize the above procedure in Algorithm \ref{main_algo} where we drop the parameters $(\mathcal{W}, \mathcal{M})$ from $\Delta \bar{L} _l^j(\mathcal{W}, \mathcal{M})$ and refer to it as $\Delta \bar{L} _l^j$ for simplicity.

\begin{algorithm}[ht]
\
    \textbf{Input:} $\mathcal{D}$ : training data, $\mathcal{W}_0$: given initialization weights, $\alpha$ : learning rate, $\mu$ : coefficient for spectral radius, $B$: bound on spectral radius, $\bar{\kappa}$: desired sparsity level,
    $N$: number of pruned neurons in each round, $E_1, E_2, E_3$: number of training epochs pre-pruning, between pruning and post-pruning.\\
    \textbf{Output:} $\mathcal{W}$ : the updated weights, $\mathcal{M}$ : the final mask vector.\\
Initialize $\mathcal{W} \leftarrow {\mathcal{W}_0}$, $\mathcal{M} \leftarrow \mathbf{e}_d$, $\kappa = 1$, $epoch = 0$.\\
\While{$\kappa > \bar{\kappa}$}{
\If{$epoch \geq E_1$ and $epoch\bmod E_2 = 0$}{
Pick a random mini-batch $\mathcal{D}^b$.\\
Compute $\Delta \bar{L} _l^j$ using \ref{dfn:equation} for mini-batch $\mathcal{D}^b$.\\
Normalize $\Delta \bar{L} _l^j$ per layer using: $\Delta\bar{L} _l^j = \frac{\Delta \bar{L} _l^j}{\sqrt{\sum_{k=1}^{n_l}(\Delta \bar{L} _l^k)^2}}$\\
Select $N$ neurons with smallest $\Delta \bar{L} _l^j$ and set their corresponding mask value in vector $\mathcal{M}_B$ equal to zero.\\
Update: $\kappa \gets \frac{\|\mathcal{W} \odot \mathcal{M}\|_0}{\|\mathcal{W} \odot e_d\|_0}$
    }
Complete one epoch of training to solve $\min_\mathcal{W} L(\mathcal{W}, \mathcal{M}_B; \mathcal{D})$ using Algorithm \ref{algo1} and update $\mathcal{W}$.\\
    Update: $epoch \gets epoch + 1$
}
Train the neural network for another $E_3$ epochs using Algorithm \ref{algo1}.
\caption{HGNP Algorithm to update parameters $\mathcal{W}$ and mask vector $\mathcal{M}_B$ for solving optimization problem \ref{dfn:our_model2}.}
\label{main_algo}
\end{algorithm}

\section{Experimental Studies}
\label{sec:computation}
In this section, we use some widely-used datasets and the HGNP algorithm to select common network models in deep learning and analyze the results. The datasets used in the  experiments are as follows. 
 
\textbf{Cifar-10}:
Collected by \cite{krizhevsky2009learning}, this dataset contains 60,000, $32\times32$ color images with 10 different labels. We use 50,000 of samples for training and keep out the remaining for validation purposes. We use conventional technique such as random horizontal flip to augment the data. 
 
\textbf{Oxford Flowers 102 dataset}: This dataset is introduced by \cite{Nilsback08} and it has 2,040 training samples and 6,149 validation samples. The number of flower categories is 102 and each image is RGB with size $224 \times 224$. We use random horizontal flip for data augmentation.

\textbf{Caltech-UCSD Birds 200 (CUB-200)}: The dataset is introduced in \cite{WelinderEtal2010} and contains 5,994 training samples and 5,794 test samples. It contains images of 200 species of birds. For the experiment on the birds dataset, we crop all the images to $160 \times 160$ to avoid memory issues while we train the model using the HGNP algorithm. 
 
We compare the quality of our pruned network and its minimum solution against Taylor \cite{molchanov2016pruning}. We chose Taylor as the only benchmark since their work is very recent and it outperforms other neuron pruning methods we mentioned in Section \ref{sec:lr}. Unless otherwise stated, we use the training data for training the model and applying the pruning scheme. In all the experiments, the accuracy on the same validation dataset is measured and is reported to compare our method with Taylor. 

We apply various common architecures AlexNet \cite{krizhevsky2012imagenet}, ResNet-18 \cite{he2016deep} and  VGG-16 \cite{simonyan2014very}. We choose Alexnet and VGG-16 as they are used in \cite{molchanov2016pruning} in their experiments. To show that our method works well with newer architectures, we select Resnet-18 as it has the power of residual networks and it can be trained in a reasonable time.We use Pytorch \cite{paszke2017automatic} for training neural networks and automatic differentiation on GPU. The Alexnet experiments are conducted using one, three Tesla K80 GPUs for the HGNP and Taylor method, respectively. All other Taylor experiments use a single GeForce RTX 2080. For the HGNP method, we use four, four and five GeForce RTX 2080 GPUs to run VGG-16/Cifar-10, ResNet-18/Cifar-10 and ResNet-18/Birds 200 dataset experiments, respectively.

\begin{figure}[ht]
\includegraphics[width=12cm]{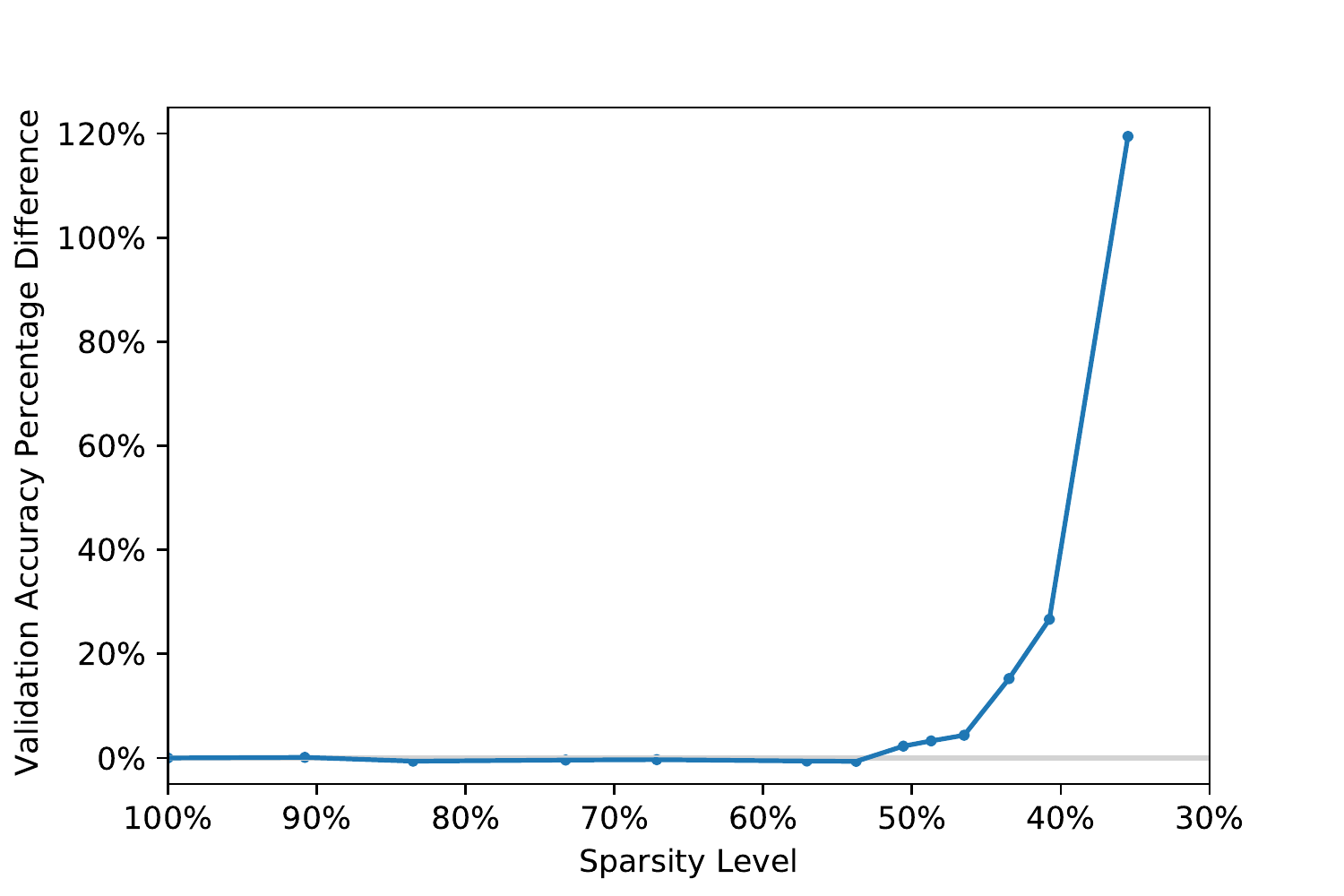}
\centering
\caption{Alexnet pruned network accuracy results on Flowers dataset.}
\label{alexnet}
\end{figure}

\subsection{Various Common Architectures}
\textbf{AlexNet experiment:} \quad For the first experiment, we follow the Taylor paper and adopt AlexNet to predict different species of flowers on the Oxford Flowers dataset. Training the network from scratch does not result in acceptable validation accuracy, so we use optimal parameters of the model trained on ImageNet \cite{imagenet_cvpr09} to initialize the network parameters except for the last fully-connected layer which is initialized randomly. We use the same hyperparameters as \cite{molchanov2016pruning} and train the model with learning rate $\alpha = 0.001$ and weight decay $0.0001$. The training data is processed by the network in mini-batch size of $32$ and we use SGD with momentum $0.9$ to update the parameters. We use $\mu = 0.001$ and $B = 0.5$ to penalize the spectral radius of Hessian. 
The hyperparameters exactly follow the values chosen by the Taylor paper. We train the network for 5 epochs initially ($E_1$) and start pruning $N=100$ neurons after each $E_2 = 5$ epochs. Once we hit the desired sparsity level, we train the network for $E_3 = 50$ epochs. 

For the Taylor method results, we use a pre-trained network which is trained for 20 epochs and prune one neuron every 30 gradient descent steps afterwards. These values have been optimized by grid search. Figure \ref{alexnet} plots the relative validation accuracy percentage difference of the HGNP and Taylor method against the sparsity level. It shows that the HGNP and Taylor methods are performing closely with less sparse models, but when the sparsity level increases, the relative gap between the accuracy of HGNP and Taylor becomes huge ($120 \%$ when both models are at $35\%$ sparsity level). For sparsity above $55\%$, the performance of both methods are very similar. For sparsity of $50\%$, $40\%$ and $35\%$ the accuracy is $(78.9\%, 71.1\%)$, $(77.0\%, 60\%)$ and $(75.5\%, 34\%)$, respectively where the first value corresponds to HGNP and the second belongs to Taylor.

\begin{figure}[ht]
\includegraphics[width=12cm]{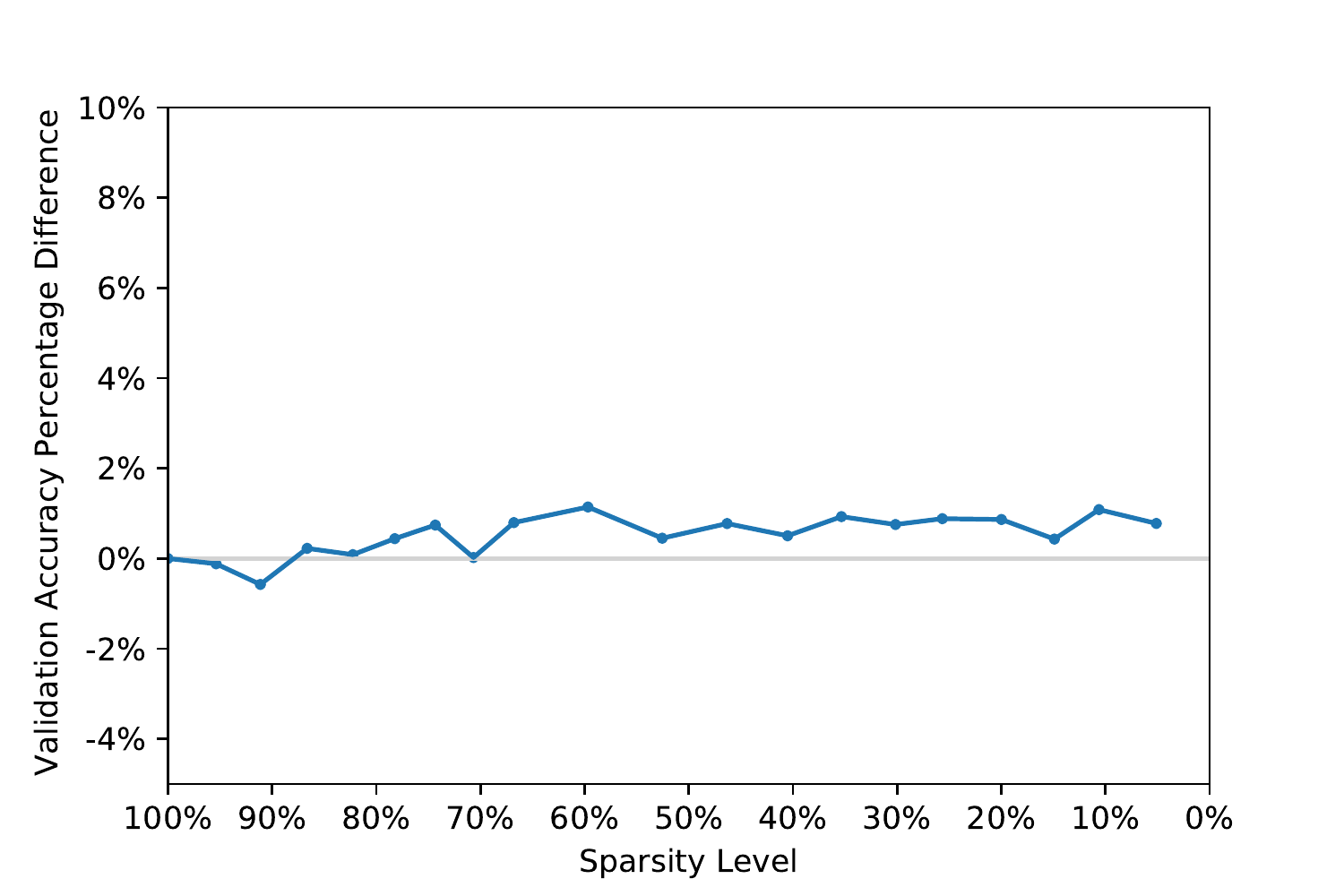}
\centering
\caption{ResNet-18 pruned network accuracy results on Cifar-10 dataset.}
\label{resnet}
\end{figure}

\textbf{ResNet-18 experiment on Cifar-10:} \quad For the second experiment, we train and prune ResNet-18 on the Cifar-10 dataset. We use mini-batch size $128$, along with weight decay $0.0005$ and initial learning rate $\alpha = 0.1$. SGD with momentum $0.9$ is used for a smoother update of the parameters. We set $\mu = 0.001$ and $B = 0.5$ for the spectral radius term. The model is trained without any pruning for $E_1 = 50$. We alternate between pruning $N=100$ neurons and training, each $E_2 = 5$ epochs, and fine-tune the model with desired sparsity level for another $E_3 = 20$ epochs. Moreover, due to the specific structure of the residual networks and existence of residual connections, some of the layers should have the same number of neurons to avoid dimension incompatibility during feed-forward. To comply with dimension compatibility, we group those specific layers together and prune the same number of neurons from each layer by averaging the number of neurons to prune suggested by our pruning method for those layers.

For the Taylor method, we train the model for $200$ epochs. Then we start pruning one neuron from the network and fine-tune the parameters with one epoch of the data. We continue alternating between pruning and fine-tuning until a desired sparsity level is reached. Figure \ref{resnet} plots the relative percentage difference in accuracy on the validation dataset between the HGNP and the Taylor methods. The HGNP method consistently outperforms at sparsity level $85\%$ and below, although the improvement is not drastic.

\textbf{ResNet-18 experiment on Birds 200 dataset:} \quad The third experiment is conducted on training ResNet-18 on the Birds 200 dataset. Like the first experiment, we rely on transfer learning and utilize the pretrained convolutional network weights of ResNet-18 on ImageNet. The learning and weight decay rates are set to $0.001$ and $0.0001$, respectively. We use mini-batch size 32 along with momentum $0.9$, $\mu= 0.001$ and $B=0.5$ for training the network. We set $E_1 = 25$ and we prune 100 neurons every 5 epochs afterwards. As our last fine tuning step, we train for $E_3=20$ more epochs after achieving the desired sparsity level.
\begin{figure}[h!]
\includegraphics[width=12cm]{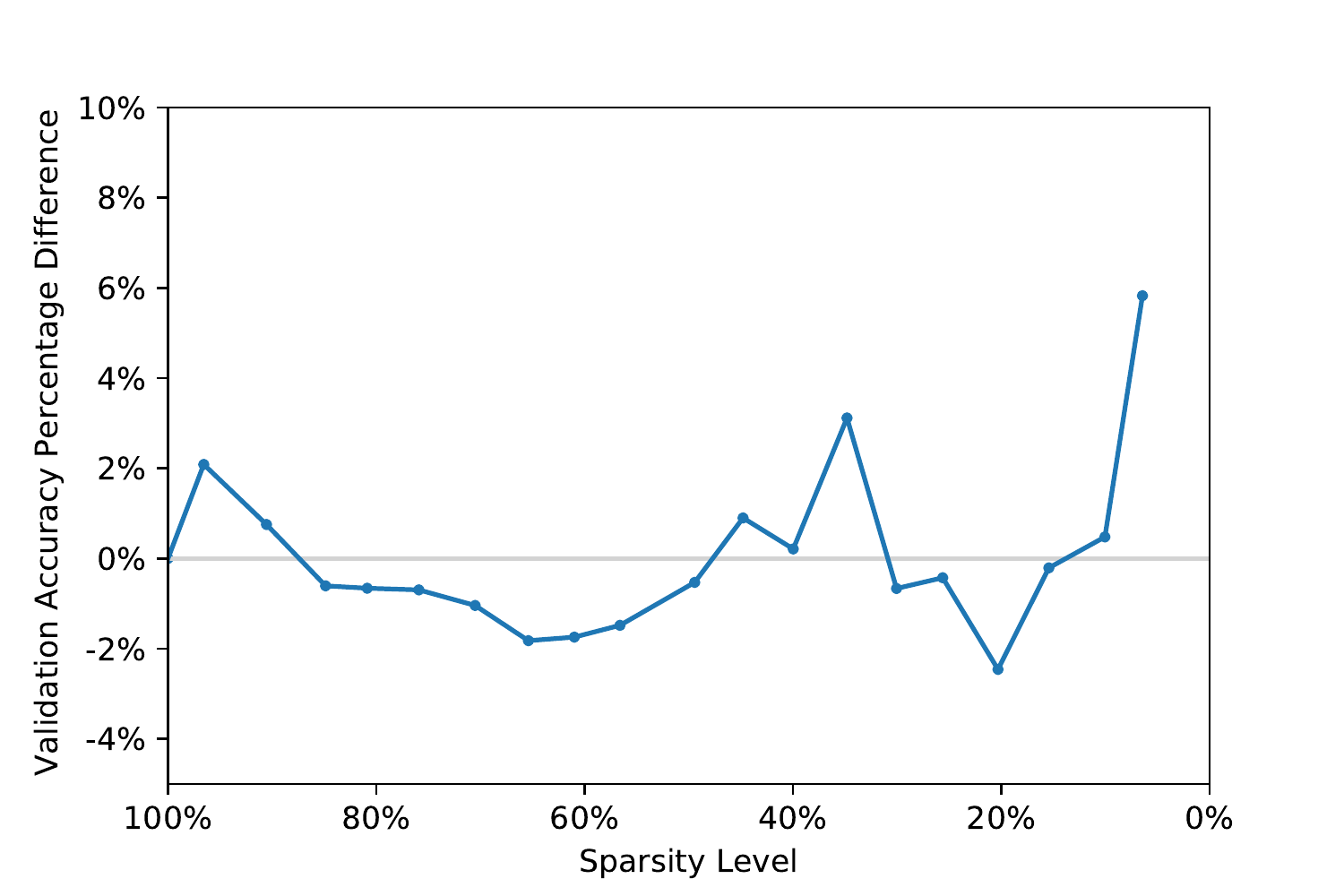}
\centering
\caption{ResNet-18 pruned network accuracy results on Birds 200 dataset.}
\label{resnet-birds}
\end{figure}
For the Taylor method, we initially train the full model for 60 epochs with learning rate $0.01$ and weight decay rate $0.0001$ for 60 epochs. The hyperparameters are optimized to achieve the best performance. We then start pruning one neuron at each pruning step and train for an epoch to recover the performance afterwards. Figure \ref{resnet-birds} suggests that our model outperforms at really high and low sparsity levels. At the sparsity level $14\%$ and below, our model starts to outperform again. At the sparsity level $5\%$, the relative accuracy gap is at $6\%$. All accuracies range from $70.5\%$ to $51.9\%$ and from $70.5\%$ to $49.0\%$ for HGNP and Taylor, respectively.

\textbf{VGG-16 experiment:} \quad Our last experiment is on training the VGG-16 neural network with batch normalization \cite{ioffe2015batch} on the Cifar-10 dataset. The Mini-batch size and the initial learning rate are set to $128$ and $0.1$, respectively. We decay the learning rate as training progresses and add a regularization term to the loss function with coefficient $0.0005$. Values $\mu$ and $B$ are set to $0.001$ and $0.5$, respectively.
\begin{figure}[ht]
\includegraphics[width=12cm]{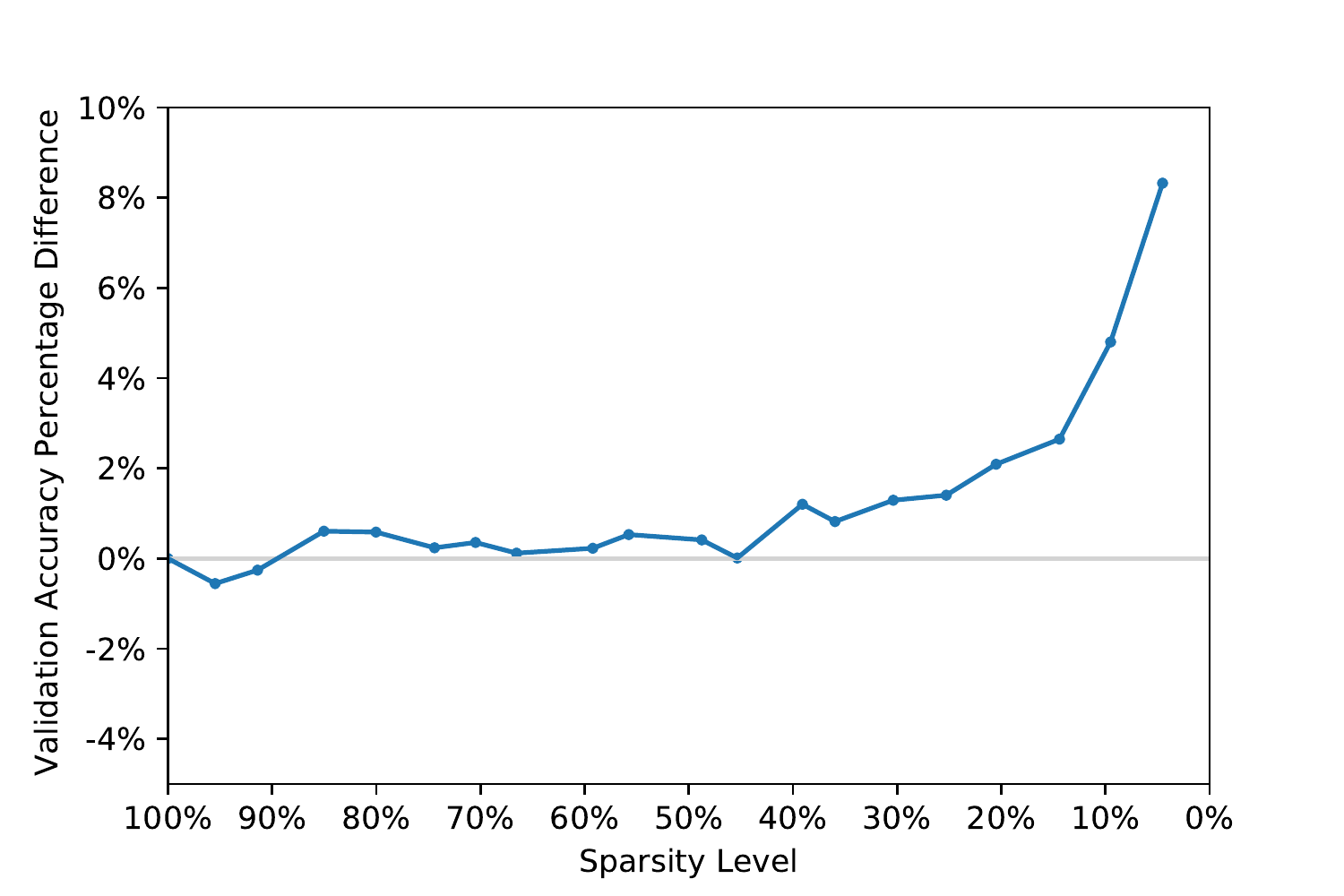}
\centering
\caption{VGG16 pruned network accuracy results on Cifar-10 dataset.}
\label{vgg}
\end{figure}
For the Taylor, we pre-train it for 200 epochs and start pruning one neuron and training for one epoch alternately. For HGNP, the pre-pruning number of training epochs is $E_1 = 50$ and we prune $N=100$ neurons at each pruning epoch and train the model for $E_2 = 5$ epochs between each pruning epoch. Finally the model is fine-tuned for another $E_3 = 20$ epochs. Figure \ref{vgg} compares the two methods. At the $5 \%$ sparsity level, the relative gap between the accuracy of HGNP and Taylor is greater than $8\%$. Our model outperforms at all sparsity levels below $90\%$. At sparsity of $10\%$, the accuracy of Taylor is $87.3\%$ while HGNP is at $91.4\%$.

\subsection{Train and Inference Times}
In this section, we compare the training and inference times of the HGNP and Taylor methods. The total training and pruning time for the HGNP method is longer than for the Taylor method as it needs to calculate the largest eigenvalue and its corresponding eigenvector for each block. The times for the experiments are summarized in Table \ref{tab:my_label}.
\begin{table}[ht]
    \centering
\small
\begin{adjustbox}{width={\textwidth},totalheight={\textheight},keepaspectratio}
\begin{tabular}{c|c|S|S|S}
\textbf{Model/Dataset} & \textbf{Hardware} & \textbf{Target Sparsity \%} & \textbf{HGNP (hours)} & \textbf{Taylor (hours)}\\ \hline
\multirow{1}{*}{AlexNet/Flowers 102} & TESLA K80 & 35.1 & 57.3 & 3.8\\
 \hline
\multirow{1}{*}{ResNet-18/Cifar-10} & GeForce RTX 2080 & 4.5  & 121.0 & 10.7\\
 \hline
\multirow{1}{*}{VGG-16/Cifar-10} & GeForce RTX 2080 & 4.5& 115.1  & 12.7\\
 \hline
 \multirow{1}{*}{ResNet-18/Birds 200 dataset} & GeForce RTX 2080 & 5.0  & 129.9 & 27.3\\
\end{tabular}
\end{adjustbox}
\caption{Comparison of the total training and pruning time across different models/dataset of HGNP vs. Taylor.}
\label{tab:my_label}
\end{table}

To compare the actual speed up for our pruned models, we measure the inference time of one mini-batch for the neuron-pruned (using HGNP), weight-pruned and full models. To calculate the inference time for weight-pruned models, we initialize the network with random weights and randomly kept a percentage (sparsity level percentage) of them and set other to zero. We believe that using optimal weights along with state-of-the-art methods in weight pruning would not have significant effects on inference time compared to the aforementioned strategy. The actual inference time can be dependent on many different parameters including the hardware. Table \ref{chap1:tab} compares the inference time for all three models using different GPUs and mini-batch sizes across various models. The full model corresponds to the $100 \%$ sparsity level for both neurons and weights. Our result shows that inference time for neuron-pruned models is lower than of the full model as expected. In addition, at the same sparsity level, neuron-pruned models are faster than weight-pruned models in feed-forward.\\
\begin{table}[ht]
  \centering
\small
\begin{adjustbox}{width={\textwidth},totalheight={\textheight},keepaspectratio}
\begin{tabular}{c|c|S|S|S|S}
\textbf{Model/Dataset} & \textbf{Hardware} & \textbf{Sparsity (Neurons) \%} & \textbf{Sparsity (Weights) \%} &\textbf{Mini-batch} & \textbf{Time(ms)} \\ \hline
\multirow{2}{*}{AlexNet/Flowers} & GPU:TESLA K80 & 100.0 & 100.0 & 5000 & 913\\
 & GPU:TESLA K80 &   & 30.4 & 5000 & 808\\
  & GPU:TESLA K80 & 30.4 &  & 5000 & 570\\ \hline
\multirow{4}{*}{ResNet-18/Cifar-10} & CPU:3.4 GHz Intel Core i5 & 100.0  & 100.0 & 128 & 3437\\
 & CPU:3.4 GHz Intel Core i5 & 4.5 &  & 128 & 1013\\
  & CPU:3.4 GHz Intel Core i5 &  & 4.5  & 128 & 2402\\
 & GPU:GeForce RTX 2080 & 100.0 & 100.0 & 2000 & 192\\
 & GPU:GeForce RTX 2080 &  & 4.5 & 2000 & 206\\
 & GPU:GeForce RTX 2080 & 4.5 &  & 2000 & 90\\ \hline
\multirow{4}{*}{VGG-16/Cifar-10} & CPU:3.4 GHz Intel Core i5 & 100.0& 100.0  & 128 & 2251\\
 & CPU:3.4 GHz Intel Core i5 & 4.5 &  & 128 & 482\\
 & CPU:3.4 GHz Intel Core i5 &  & 4.5  & 128 & 1709\\
 & GPU:GeForce RTX 2080 & 100.0 & 100.0 & 2000 & 132\\
  & GPU:GeForce RTX 2080 &  & 4.5 & 2000 & 187\\
 & GPU:GeForce RTX 2080 & 4.5 &  & 2000 & 39\\
\end{tabular}
\end{adjustbox}
\caption{Comparison of inference time for neuron-pruned, weight-pruned and full networks.}
\label{chap1:tab}
\end{table} 

We also compare neuron-pruned models with weight pruned models that produce the same level of accuracy. According to \cite{lee2018snip}, the VGG model on the Cifar-10 dataset has accuracy of $92\%$ at $3\%$ sparsity. This level of accuracy translates to $14.4\%$ sparsity in our neuron-pruned model. The inference time comparison is summarized in Table \ref{chap1:tab__}. The results show that even though the sparsity level is higher for neuron-pruned models compared to weight-pruned models to produce the same level of accuracy, neuron-pruned models are still faster in inference time.

\begin{table}[ht]
  \centering
\small
\begin{adjustbox}{width={\textwidth},totalheight={\textheight},keepaspectratio}
\begin{tabular}{c|c|S|S|S|S}
\textbf{Model/Dataset} & \textbf{Hardware} & \textbf{Sparsity (Neurons) \%} & \textbf{Sparsity (Weights) \%} &\textbf{Mini-batch} & \textbf{Time(ms)} \\ \hline
\multirow{4}{*}{VGG-16/Cifar-10} 
 & CPU:3.4 GHz Intel Core i5 & 14.4 &  & 128 & 1414\\
 & CPU:3.4 GHz Intel Core i5 &  & 3  & 128 & 1605\\
  & GPU:GeForce RTX 2080 & 14.4 &  & 2000 & 102\\
 & GPU:GeForce RTX 2080 &  & 3  & 2000 & 184\\
\end{tabular}
\end{adjustbox}
\caption{Inference time comparison of neuron-pruned and weight-pruned models that create $92\%$ accuracy.}
\label{chap1:tab__}
\end{table} 
\subsection{Ablation Study}
We conduct ablation experiments to show the effectiveness of our proposed components for pruning. To this end, we remove two components of HGNP one at a time and compare the results with the full algorithm:  1) We remove the component related to the spectral radius (set $\mu=0$) and train the model only using the cross-entropy loss. However, we use the whole loss function including the spectral radius part for deciding what neurons to prune. 2) We use the loss function with $\mu=0$ to specify which neurons to prune. However, the complete loss function is used for training and fine tuning the model.

\begin{figure}[ht]
\includegraphics[width=12cm]{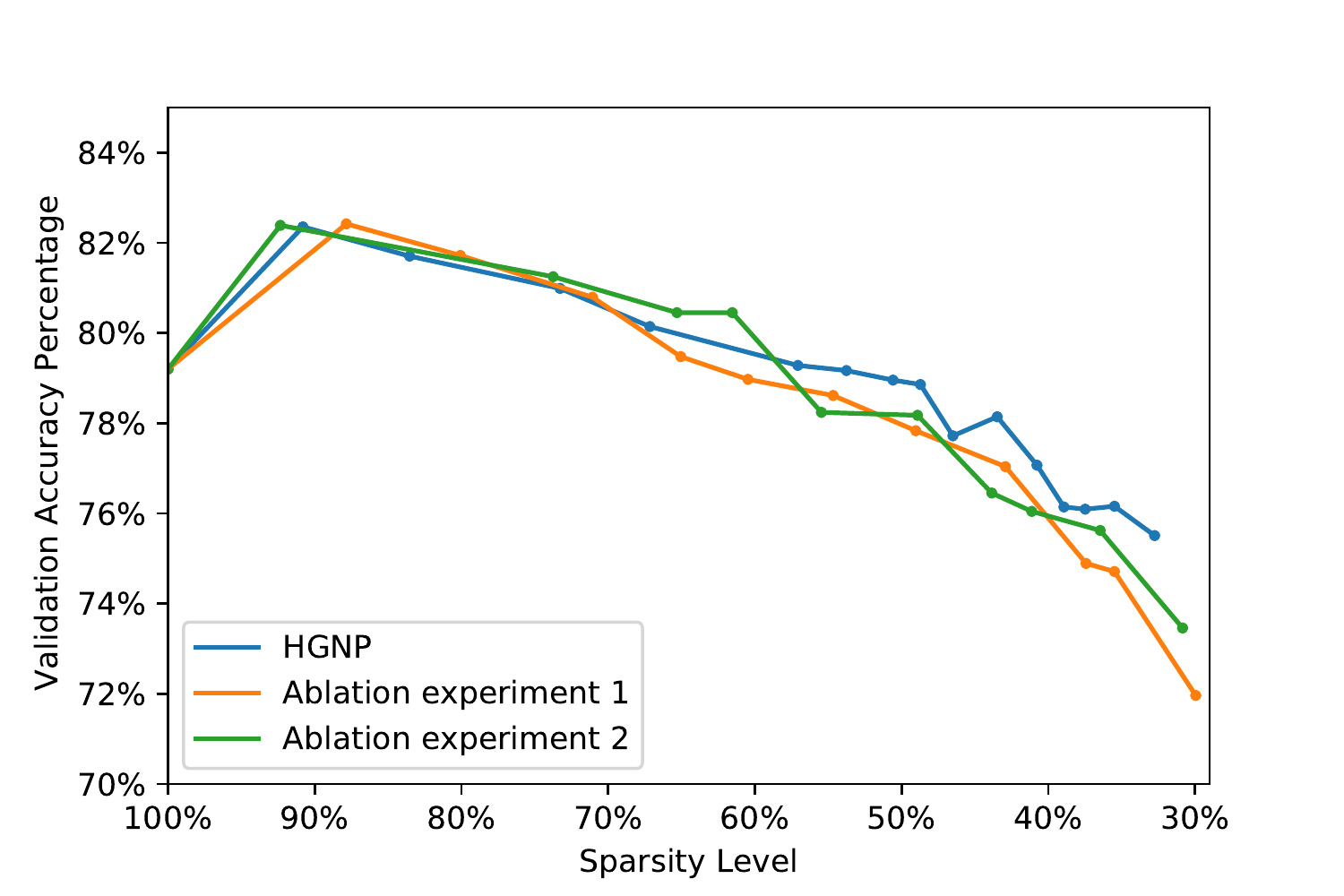}
\centering
\caption{Ablation study results for Alexnet on Flowers dataset.}
\label{ablation}
\end{figure}

We conduct ablation experiments on the AlexNet model trained on the Flowers dataset since it trains fast. The results are shown in Figure \ref{ablation}. Full HGNP is consistently better for the sparsity level less than $58\%$ which shows the effectiveness of the algorithm.

\subsection{Heuristic Method to Switch between Taylor and HGNP}
In all the experiments, the benchmark method (Taylor) works better than HGNP up to a sparsity level and beyond that our algorithm outperforms in terms of the accuracy performance. For instance, for Alexnet this threshold value is around sparsity level $91\%$ (see Figure \ref{alexnet}). If we are after an algorithm that excels for all sparsity levels, this observation leads to a strategy of using Taylor for large sparsity levels and HGNP for low levels. The challenge is identifying the switching level. To this end, we use a linear regression model with 4 samples (each one of the experiments is a sample) where the response variable is the sparsity level threshold value specifying which model outperforms. We consider two types of independent variables. The first type is related to the dataset and those in the other type characterize the neural network architecture. 
The dataset variables considered are: the number of training and validation samples, the dimension of input images, and the number of classes in the model to predict. For the neural network architecture related variables we use: the number of convolutional layers, the number of linear layers, the number of trainable parameters of the neural network model, the average kernel size across all convolutional layers, and the number of kernels. We use all these predictors and the response variable to perform Lasso regression. The best model based on $R^2$ and all p-values below $0.05$ contains only two significant predictors: the number of classes and the number of training samples. By using the model we predict the threshold sparsity level for each of the experiments and use it to decide when to switch from Taylor to HGNP. We then create accuracy plots and compare the area under the curve (AUC) for the method which switches between HGNP and Taylor (i.e. the hybrid method) versus the Taylor method in addition to HGNP versus Taylor and summarize the results in Table \ref{table_rel}. Based on the results, AUC is improved for HGNP compared to Taylor across all model/dataset pairs except for Resnet-18/Birds. We observe a substantial improvement of hybrid method compared to Taylor forAlexNet/Flowers 102 and VGG-16/Cifar-10. The hybrid method outperforms Taylor in every experiment.
\begin{table}[ht]
  \centering
\small

\begin{tabular}{ c|c|c|c }

\textbf{Model/Dataset} & \textbf{Hybrid vs Taylor} & \textbf{HGNP vs Taylor} & \textbf{Predicted Threshold}\\ \hline
AlexNet/Flowers 102 & 9.33 & 9.20 & 91.0 \\
Resnet-18/Birds  & 0.08 & -0.29 & 48.0\\
VGG-16/Cifar-10 & 1.63 & 1.59 & 87.8\\
Resnet-18/Cifar-10 & 0.47 & 0.42 & 87.8\\

\end{tabular}
\caption{Relative percentage of improvement in accuracy across different architectures/datasets.}
\label{table_rel}
\end{table}
\section{Discussion}
We propose an algorithm that prunes neurons in a DNN architecture with a special attention to the final solution of the pruned model. The algorithm is able to achieve flatter minima which generalize better on unseen data. The experiments show that the HGNP algorithm outperforms the existing literature across different neural network architectures and datasets. Our compact model achieves an actual improvement in the inference time given current software by pruning neurons instead of weights.


\bibliography{references}

\begin{thebibliography}{37}
\providecommand{\natexlab}[1]{#1}
\providecommand{\url}[1]{\texttt{#1}}
\expandafter\ifx\csname urlstyle\endcsname\relax
  \providecommand{\doi}[1]{doi: #1}\else
  \providecommand{\doi}{doi: \begingroup \urlstyle{rm}\Url}\fi

\bibitem[Amari(1998)]{amari1998natural}
S.-I. Amari.
\newblock Natural gradient works efficiently in learning.
\newblock \emph{Neural Computation}, 10\penalty0 (2):\penalty0 251--276, 1998.

\bibitem[Anwar et~al.(2017)Anwar, Hwang, and Sung]{anwar2017structured}
S.~Anwar, K.~Hwang, and W.~Sung.
\newblock Structured pruning of deep convolutional neural networks.
\newblock \emph{ACM Journal on Emerging Technologies in Computing Systems},
  13\penalty0 (3):\penalty0 1--18, 2017.

\bibitem[Botev et~al.(2017)Botev, Ritter, and Barber]{botev2017practical}
A.~Botev, H.~Ritter, and D.~Barber.
\newblock Practical {G}auss-{N}ewton optimisation for deep learning.
\newblock \emph{Journal of Machine Learning Research}, pages 557--565, 2017.

\bibitem[Chauvin(1989)]{chauvin1989back}
Y.~Chauvin.
\newblock A back-propagation algorithm with optimal use of hidden units.
\newblock In \emph{Advances in Neural Information Processing Systems}, pages
  519--526, 1989.

\bibitem[Cheng et~al.(2017)Cheng, Wang, Zhou, and Zhang]{cheng2017survey}
Y.~Cheng, D.~Wang, P.~Zhou, and T.~Zhang.
\newblock A survey of model compression and acceleration for deep neural
  networks.
\newblock \emph{arXiv preprint arXiv:1710.09282}, 2017.

\bibitem[Deng et~al.(2009)Deng, Dong, Socher, Li, Li, and
  Fei-Fei]{imagenet_cvpr09}
J.~Deng, W.~Dong, R.~Socher, L.-J. Li, K.~Li, and L.~Fei-Fei.
\newblock {ImageNet: A large-scale hierarchical image database}.
\newblock In \emph{Conference on Computer Vision and Pattern Recognition},
  2009.

\bibitem[Denil et~al.(2013)Denil, Shakibi, Dinh, Ranzato, and
  De~Freitas]{denil2013predicting}
M.~Denil, B.~Shakibi, L.~Dinh, M.~Ranzato, and N.~De~Freitas.
\newblock Predicting parameters in deep learning.
\newblock In \emph{Advances in Neural Information Processing Systems}, pages
  2148--2156, 2013.

\bibitem[Grosse and Martens(2016)]{grosse2016kronecker}
R.~Grosse and J.~Martens.
\newblock A {K}ronecker-factored approximate {F}isher matrix for convolution
  layers.
\newblock In \emph{International Conference on Machine Learning}, pages
  573--582, 2016.

\bibitem[Han et~al.(2015)Han, Pool, Tran, and Dally]{han2015learning}
S.~Han, J.~Pool, J.~Tran, and W.~Dally.
\newblock Learning both weights and connections for efficient neural network.
\newblock In \emph{Advances in Neural Information Processing Systems}, pages
  1135--1143, 2015.

\bibitem[Hassibi and Stork(1993)]{hassibi1993second}
B.~Hassibi and D.~G. Stork.
\newblock Second order derivatives for network pruning: Optimal brain surgeon.
\newblock In \emph{Advances in Neural Information Processing Systems}, pages
  164--171, 1993.

\bibitem[He et~al.(2016)He, Zhang, Ren, and Sun]{he2016deep}
K.~He, X.~Zhang, S.~Ren, and J.~Sun.
\newblock Deep residual learning for image recognition.
\newblock In \emph{Proceedings of the IEEE Conference on Computer Vision and
  Pattern Recognition}, pages 770--778, 2016.

\bibitem[Hu et~al.(2016)Hu, Peng, Tai, and Tang]{hu2016network}
H.~Hu, R.~Peng, Y.-W. Tai, and C.-K. Tang.
\newblock Network trimming: A data-driven neuron pruning approach towards
  efficient deep architectures.
\newblock \emph{arXiv preprint arXiv:1607.03250}, 2016.

\bibitem[Ioffe and Szegedy(2015)]{ioffe2015batch}
S.~Ioffe and C.~Szegedy.
\newblock Batch normalization: Accelerating deep network training by reducing
  internal covariate shift.
\newblock \emph{International Conference on Machine Learning}, pages 448--456,
  2015.

\bibitem[Ishikawa(1996)]{ishikawa1996structural}
M.~Ishikawa.
\newblock Structural learning with forgetting.
\newblock \emph{Neural Networks}, 9\penalty0 (3):\penalty0 509--521, 1996.

\bibitem[Jastrzebski et~al.(2018)Jastrzebski, Kenton, Arpit, Ballas, Fischer,
  Bengio, and Storkey]{jastrzebski2018finding}
S.~Jastrzebski, Z.~Kenton, D.~Arpit, N.~Ballas, A.~Fischer, Y.~Bengio, and
  A.~J. Storkey.
\newblock Finding flatter minima with sgd.
\newblock In \emph{International Conference on Learning Representations
  (Workshop)}, 2018.

\bibitem[Keskar et~al.(2017)Keskar, Nocedal, Tang, Mudigere, and
  Smelyanskiy]{keskar2016large}
N.~S. Keskar, J.~Nocedal, P.~T.~P. Tang, D.~Mudigere, and M.~Smelyanskiy.
\newblock On large-batch training for deep learning: Generalization gap and
  sharp minima.
\newblock In \emph{5th International Conference on Learning Representations},
  2017.

\bibitem[Kingma and Ba(2014)]{kingma2014adam}
D.~P. Kingma and J.~Ba.
\newblock Adam: A method for stochastic optimization.
\newblock \emph{arXiv preprint arXiv:1412.6980}, 2014.

\bibitem[Krizhevsky et~al.(2012)Krizhevsky, Sutskever, and
  Hinton]{krizhevsky2012imagenet}
A.~Krizhevsky, I.~Sutskever, and G.~E. Hinton.
\newblock Imagenet classification with deep convolutional neural networks.
\newblock In \emph{Advances in Neural Information Processing Systems}, pages
  1097--1105, 2012.

\bibitem[Krizhevsky et~al.(2009)]{krizhevsky2009learning}
A.~Krizhevsky et~al.
\newblock Learning multiple layers of features from tiny images.
\newblock \emph{Master's thesis, University of Toronto}, 2009.

\bibitem[LeCun et~al.(1990)LeCun, Denker, and Solla]{lecun1990optimal}
Y.~LeCun, J.~S. Denker, and S.~A. Solla.
\newblock Optimal brain damage.
\newblock \emph{Advances in Neural Information Processing Systems}, pages
  598--605, 1990.

\bibitem[LeCun et~al.(2012)LeCun, Bottou, Orr, and
  M{\"u}ller]{lecun2012efficient}
Y.~LeCun, L.~Bottou, G.~B. Orr, and K.-R. M{\"u}ller.
\newblock Efficient backprop.
\newblock In \emph{Neural networks: Tricks of the Trade}, pages 9--48.
  Springer, 2012.

\bibitem[Lee et~al.(2018)Lee, Ajanthan, and Torr]{lee2018snip}
N.~Lee, T.~Ajanthan, and P.~H. Torr.
\newblock Snip: Single-shot network pruning based on connection sensitivity.
\newblock \emph{arXiv preprint arXiv:1810.02340}, 2018.

\bibitem[Li et~al.(2016)Li, Kadav, Durdanovic, Samet, and Graf]{li2016pruning}
H.~Li, A.~Kadav, I.~Durdanovic, H.~Samet, and H.~P. Graf.
\newblock Pruning filters for efficient convnets.
\newblock \emph{arXiv preprint arXiv:1608.08710}, 2016.

\bibitem[Martens(2020)]{martens2014new}
J.~Martens.
\newblock New insights and perspectives on the natural gradient method.
\newblock \emph{Journal of Machine Learning Research}, 21:\penalty0 1--76,
  2020.

\bibitem[Martens and Grosse(2015)]{martens2015optimizing}
J.~Martens and R.~Grosse.
\newblock Optimizing neural networks with {K}ronecker-factored approximate
  curvature.
\newblock In \emph{International Conference on Machine Learning}, pages
  2408--2417, 2015.

\bibitem[Molchanov et~al.(2016)Molchanov, Tyree, Karras, Aila, and
  Kautz]{molchanov2016pruning}
P.~Molchanov, S.~Tyree, T.~Karras, T.~Aila, and J.~Kautz.
\newblock Pruning convolutional neural networks for resource efficient
  inference.
\newblock \emph{arXiv preprint arXiv:1611.06440}, 2016.

\bibitem[Nilsback and Zisserman(2008)]{Nilsback08}
M.-E. Nilsback and A.~Zisserman.
\newblock Automated flower classification over a large number of classes.
\newblock In \emph{Indian Conference on Computer Vision, Graphics and Image
  Processing}, 2008.

\bibitem[Pascanu and Bengio(2013)]{pascanu2013revisiting}
R.~Pascanu and Y.~Bengio.
\newblock Revisiting natural gradient for deep networks.
\newblock \emph{arXiv preprint arXiv:1301.3584}, 2013.

\bibitem[Paszke et~al.(2017)Paszke, Gross, Chintala, Chanan, Yang, DeVito, Lin,
  Desmaison, Antiga, and Lerer]{paszke2017automatic}
A.~Paszke, S.~Gross, S.~Chintala, G.~Chanan, E.~Yang, Z.~DeVito, Z.~Lin,
  A.~Desmaison, L.~Antiga, and A.~Lerer.
\newblock Automatic differentiation in pytorch.
\newblock 2017.

\bibitem[Pearlmutter(1994)]{pearlmutter1994fast}
B.~A. Pearlmutter.
\newblock Fast exact multiplication by the hessian.
\newblock \emph{Neural Computation}, 6\penalty0 (1):\penalty0 147--160, 1994.

\bibitem[Sandler et~al.(2021)Sandler, Klabjan, and Luo]{adam}
A.~Sandler, D.~Klabjan, and Y.~Luo.
\newblock Non-convex optimization with spectral radius regularization.
\newblock \emph{arXiv preprint arXiv:2102.11210}, 2021.

\bibitem[Schacke(2004)]{schacke2004kronecker}
K.~Schacke.
\newblock On the {K}ronecker product.
\newblock \emph{Master's thesis, University of Waterloo}, 2004.

\bibitem[Simonyan and Zisserman(2014)]{simonyan2014very}
K.~Simonyan and A.~Zisserman.
\newblock Very deep convolutional networks for large-scale image recognition.
\newblock \emph{arXiv preprint arXiv:1409.1556}, 2014.

\bibitem[Singh et~al.(2019)Singh, Verma, Rai, and Namboodiri]{singh2019play}
P.~Singh, V.~K. Verma, P.~Rai, and V.~P. Namboodiri.
\newblock Play and prune: Adaptive filter pruning for deep model compression.
\newblock In \emph{28th International Joint Conference on Artificial
  Intelligence}, pages 3460--3466, 2019.

\bibitem[Srinivas and Babu(2015)]{srinivas2015data}
S.~Srinivas and R.~V. Babu.
\newblock Data-free parameter pruning for deep neural networks.
\newblock \emph{arXiv preprint arXiv:1507.06149}, 2015.

\bibitem[Tieleman and Hinton(2012)]{hinton2012neural}
T.~Tieleman and G.~Hinton.
\newblock Lecture 6.5-rmsprop: Divide the gradient by a running average of its
  recent magnitude.
\newblock \emph{Coursera: Neural Networks for Machine Learning}, 4\penalty0
  (2):\penalty0 26--31, 2012.

\bibitem[Welinder et~al.(2010)Welinder, Branson, Mita, Wah, Schroff, Belongie,
  and Perona]{WelinderEtal2010}
P.~Welinder, S.~Branson, T.~Mita, C.~Wah, F.~Schroff, S.~Belongie, and
  P.~Perona.
\newblock {Caltech-UCSD Birds 200}.
\newblock Technical Report CNS-TR-2010-001, California Institute of Technology,
  2010.

\end{thebibliography}

\end{document}